\documentclass{mva_style}
\usepackage{graphicx}
\usepackage{cite}
\usepackage{subcaption}
\usepackage{array}
\usepackage{float}
\usepackage{amsmath}
\usepackage{amssymb}
\usepackage[]{algorithm2e}
\usepackage[colorinlistoftodos]{todonotes}
\usepackage{multicol}

\finalcopy 

\begin{document}
\title{A Three-Player GAN: Generating Hard Samples To Improve Classification Networks}
\author{
  Simon Vandenhende, Bert De Brabandere, Davy Neven and Luc Van Gool\\
  KU Leuven\\
  ESAT-PSI, Belgium \\
  {\tt firstname.lastname@esat.kuleuven.be}\\
}

\maketitle

\section*{\centering Abstract}
\textit{
We propose a Three-Player Generative Adversarial Network to improve classification networks. In addition to the game played between the discriminator and generator, a competition is introduced between the generator and the classifier. The generator's objective is to synthesize samples that are both realistic and hard to label for the classifier.  Even though we make no assumptions on the type of augmentations to learn, we find that the model is able to synthesize realistically looking examples that are hard for the classification model. Furthermore, the classifier becomes more robust when trained on these difficult samples. The method is evaluated on a public dataset for traffic sign recognition.}

\section{Introduction}
\label{sec: introduction}
Deep convolutional neural networks have brought significant progress to the area of computer vision. However, training the models still requires vast amounts of data. As intelligent vision systems are being deployed in increasingly dynamic environments, collecting the necessary data becomes a tedious task. 

Recent work in generative modeling, based on Generative Adversarial Networks (GANs) \cite{goodfellow2014generative,miyato2018spectral,zhang2018self,brock2018large}, allows to efficiently synthesize novel samples that belong to the data distribution. GANs derive the data distribution from an adversarial game, played between two entities: the generator $G$ synthesizes new samples, and the discriminator $D$ tries to separate real samples from the ones synthesized by $G$. The goal of the generator is to confuse $D$ so that it cannot discriminate between real and fake examples. The game ends when the two players are at a Nash equilibrium. 

GANs prove useful to improve the performance of classification networks. For example, \cite{peng2018jointly} proposes an adversarial approach which jointly optimizes the data augmentation and a network for pose estimation. The generator learns to synthesize augmentations from the training data that are hard to label for the classification network. The augmentations are composed of rotations, scaling transformations and occlusions. 

Furthermore, \cite{springenberg2015unsupervised,odena2016semi,salimans2016improved,li2017triple} have successfully employed GANs in a semi-supervised learning setting. In \cite{springenberg2015unsupervised}, the discriminator learns the classification task from unlabeled data. The discriminator has to classify each sample into a chosen number of categories. Since the conditional distribution $p\left(c|x\right)$ is unknown, a goodness of fit measure is included to ensure correspondence between the categories and the class labels. \cite{salimans2016improved} trains a classifier in a semi-supervised manner by considering images from the GAN as samples from an additional class. \cite{li2017triple} trains the classifier and the generative model simultaneously. They find that both generator and classifier represent a conditional distribution between labels and images. This observation leads to a compatibility criterion between the generator and classifier.

Our work implements a three-player adversarial game in which the classification network participates. The generator adapts itself to both the discriminator and classifier. This allows the generator to estimate the distribution of samples that are hard to label correctly for the classifier. In contrast to \cite{peng2018jointly}, our work does not restrict the type of augmentations that can be learned. Also, the proposed method simply relies on backpropagation, which makes it a very general approach. We show that the three-player game can improve classification networks, when annotated data is scarce. The proposed method is evaluated on CURE-TSR \cite{temel2017cure}, a publicly available dataset for traffic sign recognition.

\section{Method}
\label{sec: method}
A regular Generative Adversarial Network \cite{goodfellow2014generative} comprises a min-max game, played between the discriminator $D$ and generator $G$. Additionally, we now introduce a competition between the generator and classifier. The objective for $G$ changes from synthesizing images that are realistic, to generating images that are both realistic and challenging for the classification network. 

As before, the discriminator is trained to predict whether a sample is real or fake. The generator, in turn, optimizes the sum of two losses. The first term is the regular GAN loss, provided by the discriminator. In order for the generator to compete with the classifier, the second loss term needs to be chosen appropriately. To this end, backpropagation should yield the maximization of the classification model's loss, on samples from $G$. This encourages $G$ to move towards the distribution of samples that confuse the classifier. The classifier is trained by minimizing the classification loss on samples from $G$. The game is played by updating all three models one after another. 

\begin{figure}
	\centering
	\includegraphics[width=0.90\linewidth]{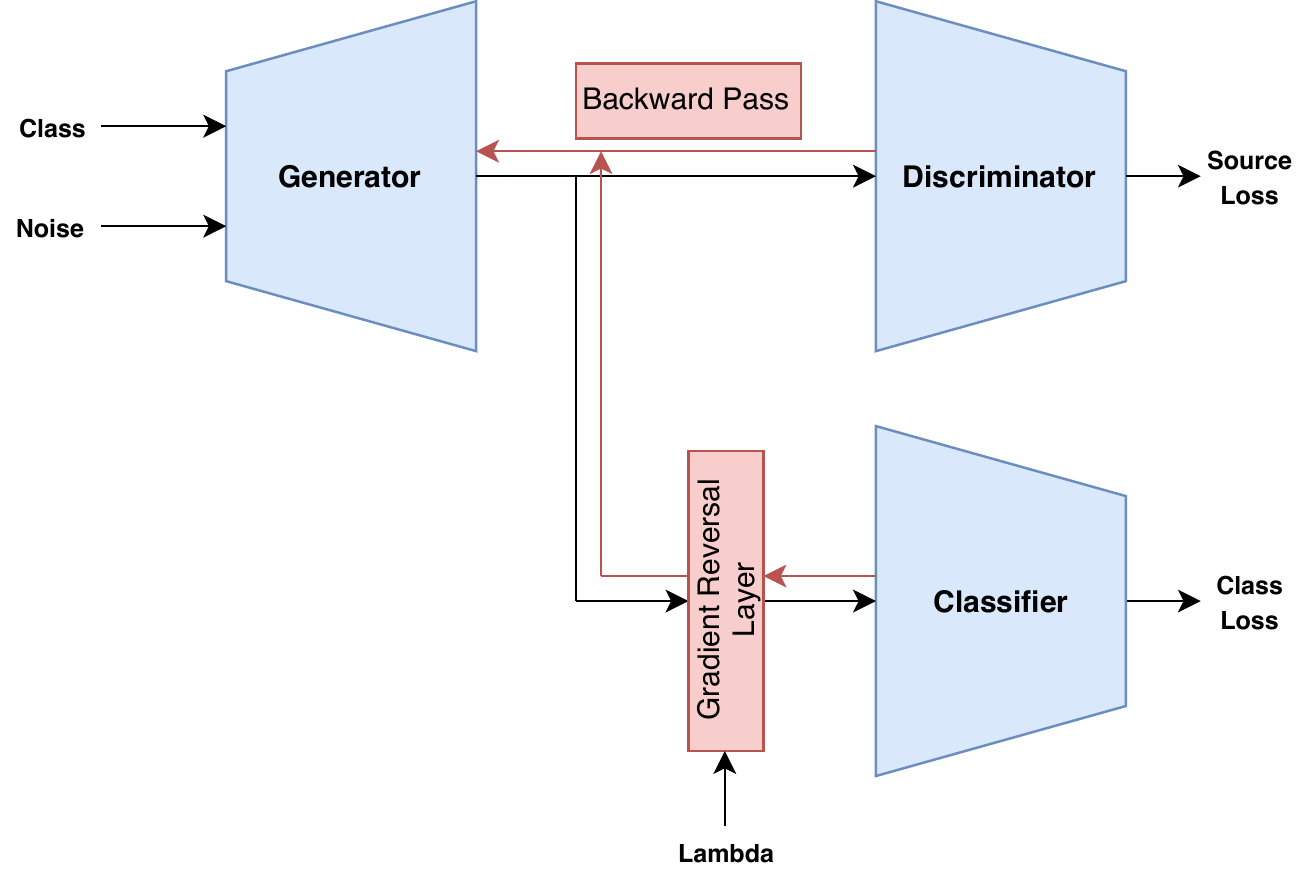}
	\caption[Training With Difficult Samples - Setup]{Setup for the three-player game. Images from the generator are propagated through both the discriminator $D$ and the classifier $C$. The gradient that is backpropagated through $D$ proceeds as usual. The gradient that is backpropagated through $C$ is rescaled and inverted as $- \lambda \nabla_{\theta_C} L_C$. The loss from $D$ penalizes $G$ for synthesizing unrealistic samples, while the inverted loss from $C$ rewards $G$ for synthesizing difficult samples.}
	\label{fig: setup}
\end{figure}
Inspired by \cite{ganin2014unsupervised}, the objective for the second loss term, seen by $G$, is realized by implementing a gradient reversal layer between the generator and classifier. During the forward pass, samples from $G$ are simply passed to the classification network. When backpropagating the classification loss, the sign of the gradient is reversed, causing the update in $G$ to maximize the classification loss. This technique is related to \cite{goodfellow2015explaining}, which finds adversarial examples by applying perturbations that lie along directions where the classification loss is likely to increase. The setup of our system is shown in figure \ref{fig: setup}. 

The three-player GAN shows some similarities with auxiliary classifier GANs (ACGANs) \cite{odena2017conditional}. In the ACGAN model, the discriminator categorizes the images in addition to predicting their source. This allows the discriminator to be deployed as a classification model. There are two main differences with our approach. First, in the three-player game, the generator tries to maximize the classification loss rather than minimize it. The focus of this work is on the generation of hard samples. Secondly, the three-player GAN separates the network architecture of the discriminator and classifier. This allows to specialize the architecture of the discriminator and classifier for their respective tasks. 

The complete training procedure for the three-player game is defined in algorithm \ref{alg: algorithm}. A hyperparameter $\lambda$ is introduced to weigh the classification loss against the discriminative loss.

\RestyleAlgo{boxruled}
{\SetAlgoNoLine
\begin{algorithm}[t]
	\caption{The three-player GAN}
	\label{alg: algorithm}
	\For{number of training iterations}{
    \textbullet Sample a batch $\left(x_g, y_g\right)$ of size $m$ from the generator, and a batch $\left(x,y\right)$ of size $m$ from the training data. \\
	\textbullet Update the discriminator by ascending its stochastic gradient: 
	\begin{equation*}
	\begin{split}
    \nabla_{\theta_d} & \Biggl[ \frac{1}{m} \sum_{(x,y)}  \log D \left(x, y\right)\\
    & + \frac{1}{m} \sum_{(x_g,y_g)} \log \left(1-D\left(x_g,y_g\right) \right) \Biggr]
    \end{split}
    \end{equation*}
    
	\textbullet Sample a batch $\left(x_g,y_g\right)$ of size $m$ from the generator. \\ 
	\textbullet Update the generator by descending its stochastic gradient:
	\begin{equation*}
	\begin{split}
	\nabla_{\theta_g} & \Biggl[ \frac{1}{m} \sum_{\left(x_g, y_g\right)} \log \left(1 - D\left(x_g, y_g\right)\right) \\
    & -\lambda \nabla_{\theta_c}  \left( \frac{1}{m} \sum_{\left(x_g,y_g\right)}^{} L_C \left(x_g, C\left(y_g\right) \right) \right) \Biggr]
    \end{split}
  	\end{equation*} \\
	\textbullet Sample a batch $\left(x,y\right)$ of size $m$. \\
	\textbullet Update the classifier by descending its stochastic gradient: 
	\begin{equation*}
		\nabla_{\theta_c} \left[ \frac{1}{m} \sum_{\left(x, y\right)} L_c\left(x, C\left(y\right) \right) \right]
	\end{equation*}
}
\end{algorithm}}

\section{Experiments}
\label{sec: experiments}
We first consider a toy example, which demonstrates that the three-player game acts as a regularizer for the decision surface of the classifier. In the second part we evaluate our method on CURE-TSR \cite{temel2017cure}. Both experiments compare the performance of a classification network trained through the three-player game against several other training scenarios. 

\begin{figure*}[ht]
\centering
\subcaptionbox{Trained on real samples.}{\includegraphics[width=0.31\textwidth]{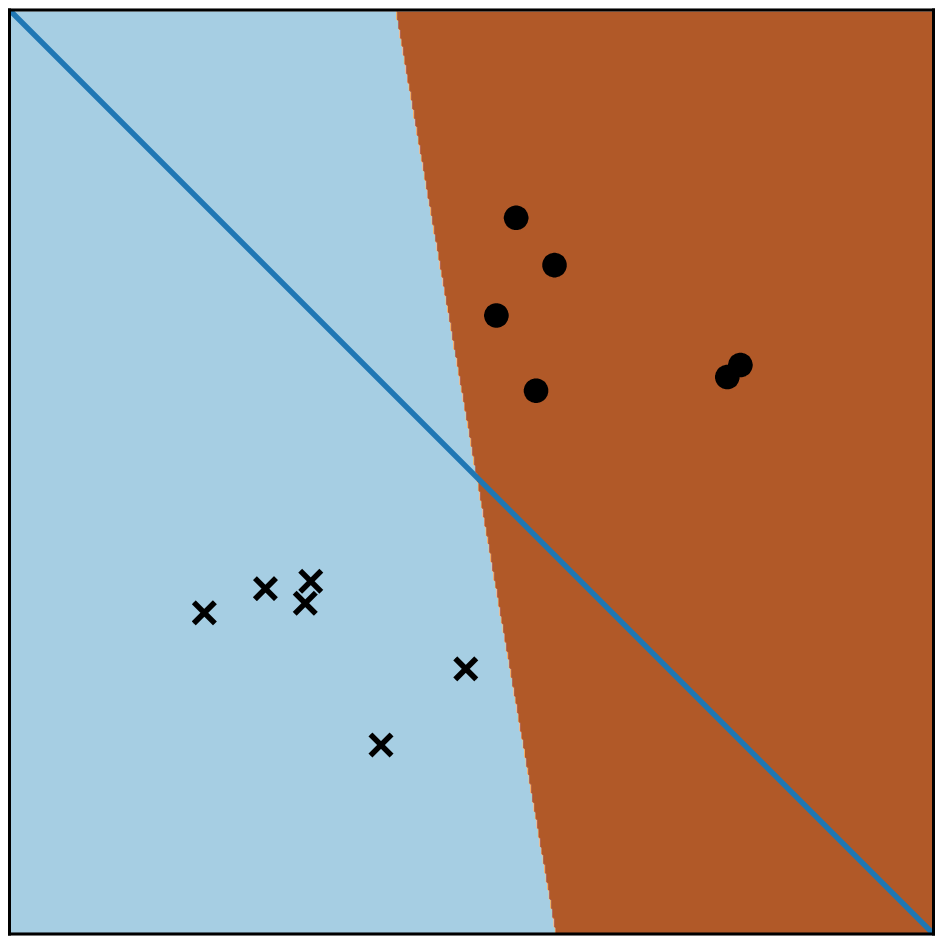}}%
\hfill
\subcaptionbox{Trained on real and synthesized samples.}{\includegraphics[width=0.31\textwidth]{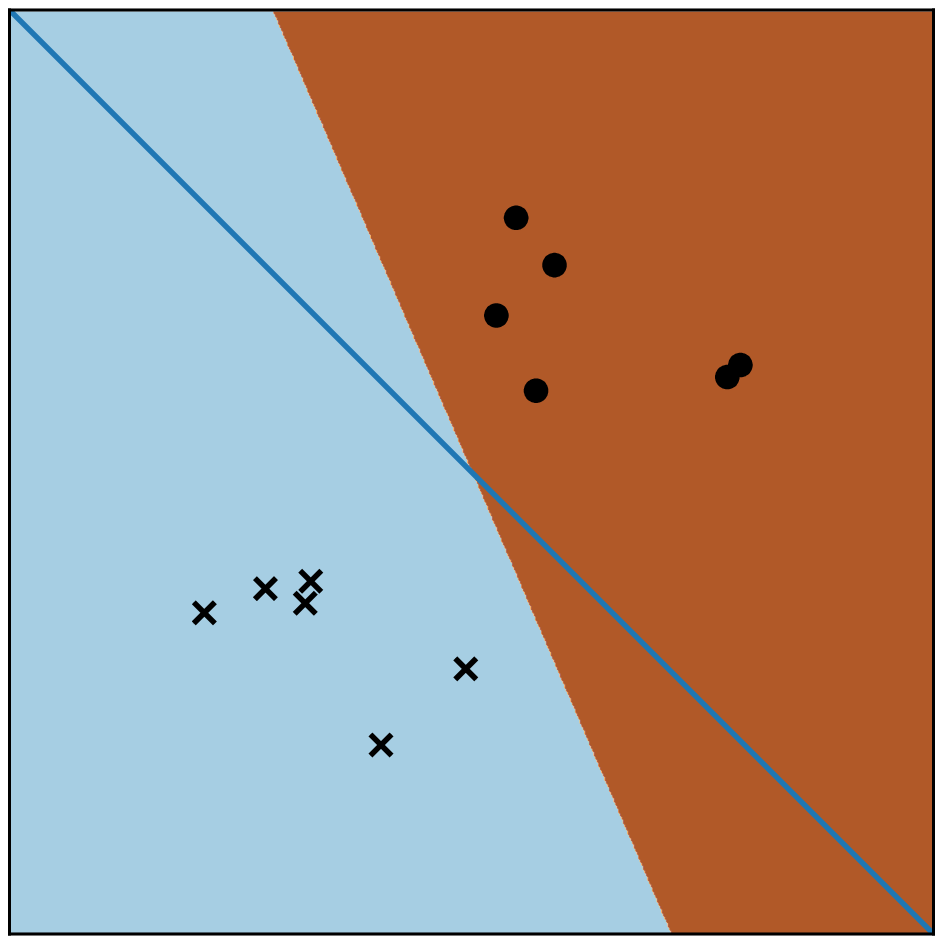}}%
\hfill
\subcaptionbox{Trained on real, synthesized and difficult samples.}{\includegraphics[width=0.31\textwidth]{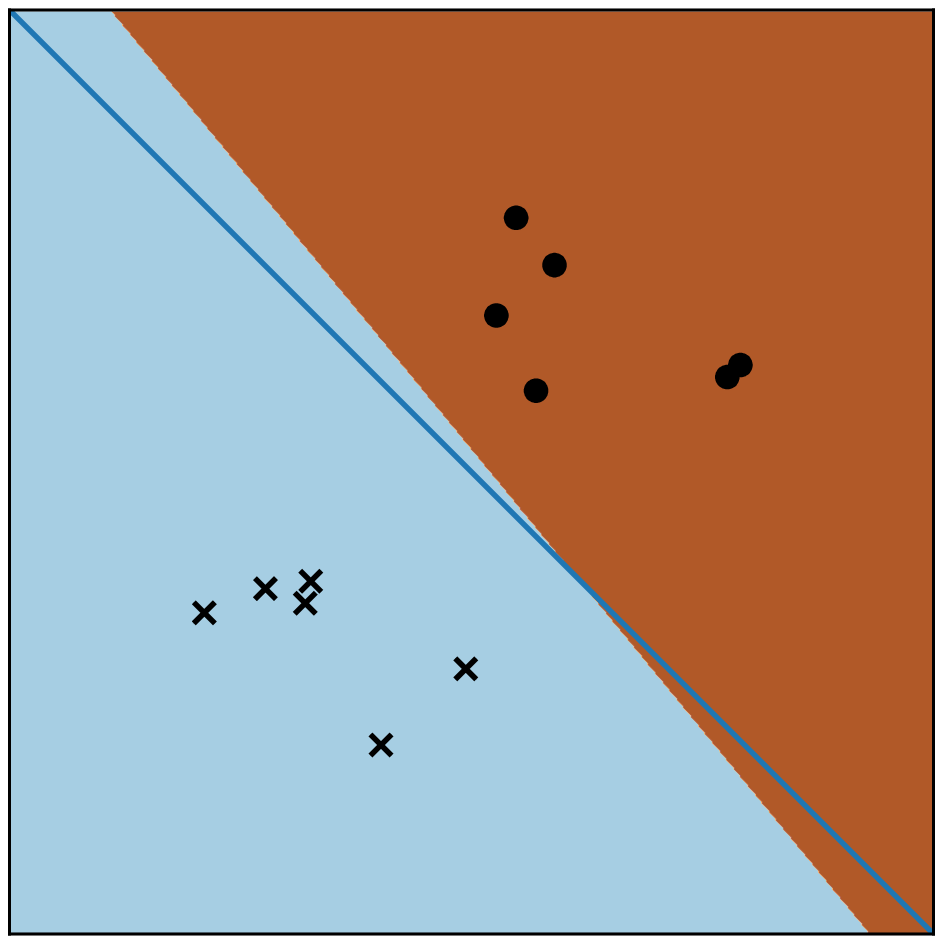}}%
\caption{Decision surface of the classification model at the end of different training procedures.}
\label{fig: toy_decision}
\end{figure*}

\subsection{Training details}
\label{sec: experiments_1}
We initialize the discriminator and generator in the three-player game by   training a conditional GAN. When updating the classifier, we sample batches containing both real images, images synthesized by the initial generator and images synthesized by the current generator. The samples from the initial generator serve to avoid catastrophic forgetting of examples that are difficult early on. 

The learning rates and the weighing parameter $\lambda$ are updated according to the scheme from \cite{springenberg2015unsupervised},
\begin{equation*}
\lambda = \frac{2 \cdot w_c}{1 + \exp \left( -10 \cdot p \right)} - 1, \mu = \frac{\mu_0}{ \left(1 + \alpha \cdot p \right)^\beta }    
\end{equation*}
with $p$ the training progress growing linearly from 0 to 1, $\alpha = 10$, $\beta = 0.75$, $w_c = 0.1$ and $\mu_0$ the initial learning rate. The value of $w_c$ is chosen smaller than one to ensure that synthesizing realistic samples has priority over synthesizing difficult ones. The weighing parameter $\lambda$ gradually grows during training, allowing the generator to come up with difficult samples even when the classification model becomes better.

\subsection{Toy example}
We demonstrate that the three-player GAN effectively acts as a regularizer, by means of a toy example. Consider the case where samples from two classes need to be separated. Both classes are distributed as two-dimensional Gaussians, parameterized by $\mu_X = \pm 1, \mu_Y=\pm 1 $ and  $\sigma_X=\sigma_Y=0.5$. The training data consists of eight examples per class, drawn as dots and crosses in figure \ref{fig: toy_decision}. The classifier, represented as a simple linear mapping, is trained using a hinge loss.

A baseline classifier and conditional GAN are trained using the available training examples. A second classification model is trained on a combination of real and synthesized samples. Thirdly, we also train a classification model based on the three-player game. For this particular example, we initialize the classifier as the baseline model and freeze its parameters. The game is played for a few epochs, allowing the generator to estimate the distribution of samples that are difficult for the baseline model. The parameters of the classification model were initialized randomly. To ensure a fair comparison, we made sure that the classification model uses the same initial weights. Figure \ref{fig: toy_decision} shows the decision boundary of the classification models obtained by different training schemes. Through comparison we find that the three-player GAN is able to regularize the decision surface. 

\begin{figure}[H]
\centering
\begin{subfigure}[b]{1.0\linewidth}
    \centering
    \includegraphics[width=0.8\linewidth]{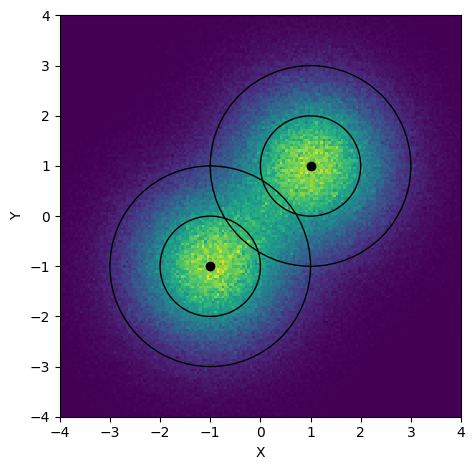}
    \caption{The true distribution consists of two classes, both distributed as two-dimensional Gaussians.}
\end{subfigure}

\begin{subfigure}[b]{1.0\linewidth}
    \centering
    \includegraphics[width=0.8\linewidth]{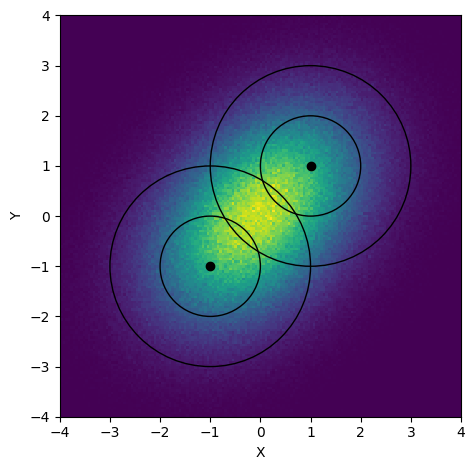}
    \caption{Distribution learned by generator during the three-player game. The generator synthesizes samples located in the area where the two class distributions overlap.}
\end{subfigure}
\caption{The real distribution consists of two classes which partially overlap. Both are two-dimensional Gaussians of which the mean is indicated as a dot. The circles in the figure correspond to multiples of the standard deviation. We find that the three-player GAN learns to synthesize samples at locations where the two classes overlap. These are  samples which are hard to label correctly for a classification model.}
\label{fig: toy_sampling}
\end{figure}

Consider again the two Gaussian distributions from before, but with an increased variance. When sampling from the two classes, we find that the distributions show a significant overlap near the origin. If the three-player game behaves as intended, we expect the generator to synthesize samples which lie near the origin. We train a classification model and a conditional GAN by sampling from the two Gaussian distributions. Afterwards, the generator is updated through the three-player game in order to synthesize difficult samples. Figure \ref{fig: toy_sampling} shows the results. We find that the generator learns to synthesize samples at locations where the classifier has a hard time. 

\subsection{CURE-TSR}
The CURE-TSR dataset \cite{temel2017cure} is composed of both real and simulated images of 14 traffic sign classes under various weather conditions. The set of simulated traffic sign instances is considered here under the following conditions: clear weather, low-mid-high levels of snow, low-mid-high levels of rain and low-mid-high levels of dark weather. For the training (resp. validation) set we took the first 100 (resp. last 50) images per class from each weather condition. Since the data contains sequences of images for which the camera gradually moves closer to the traffic sign, the data selection comes down to using only a few of such sequences.

Again, we train a classifier and conditional GAN using the available training data. A second classification model is trained using both real samples and samples synthesized by the conditional GAN. Thirdly, we compare with an auxiliary classifier GAN. Finally, a classifier is learned by means of the three-player game. As mentioned in section \ref{sec: experiments_1}, the discriminator and generator are initialized as the models from the conditional GAN. 

\begin{figure}
    \centering
    \includegraphics[width=0.9\linewidth]{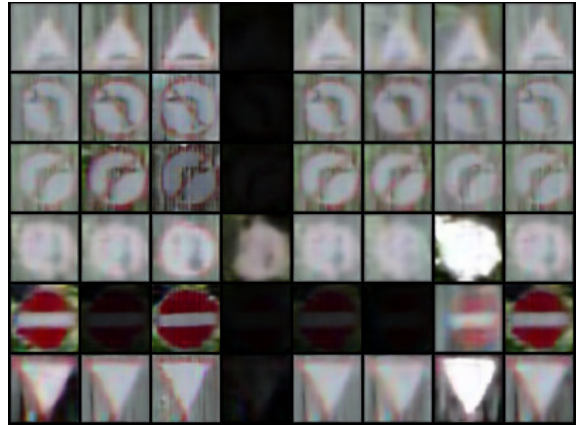}
    \caption{Images generated during the three-player game.}
    \label{fig: difficult}
\end{figure}

The network architecture for the classifier is based on one column from the multi-column deep neural network used for traffic sign recognition in \cite{cirecsan2012multi}. The discriminator and generator networks are based on earlier work \cite{miyato2018spectral}. More details can be found in the supplemental materials. The classification network was trained for 150 epochs with an Adam Optimizer \cite{kingma2014adam} $ \left(\mu_0 = 0.001, \beta_1 = 0.5, \beta_2 = 0.999 \right)$. The learning rate is degraded by a factor 10 every 60 epochs. A weight decay term of $1e-4$ is included in the classification loss. The conditional GAN was trained for 500 epochs using batches of size 64. For the auxiliary classifier GAN, we reused the architecture and training scheme from the original work \cite{odena2017conditional}. We used an Adam Optimizer with the same learning parameters as \cite{miyato2018cgans} $\left(\mu_0 = 0.0002, \beta_1 = 0.0, \beta_2 = 0.9 \right)$. The initial learning rates for the three-player game are the same as for the other training strategies. The results can be found in table \ref{tab: cure-tsr}. We find that training the classification model by means of the three-player game improves the test accuracy. Figure \ref{fig: difficult} shows images that were generated during the three-player game. 

\begin{table}[H]
\caption{Accuracy on the CURE-TSR test set for different training schemes.}
\begin{center}
    \begin{tabular}{|l|c|c|}
    \hline
    \makebox[20mm]{Method} & \makebox[30mm]{Without real data} & \makebox[20mm]{With real data} \\
    \hline
    Baseline    &   -           & 83.38          \\
    cGAN        & 77.08         & 83.23          \\
    ACGAN       & -             & 79.23          \\
    Threeplayer & \textbf{79.83}& \textbf{85.41} \\
    \hline
    \end{tabular}
    \label{tab: cure-tsr}
\end{center}
\end{table}

\section{Conclusion}
\label{sec: conclusion}
We have proposed an effective, yet simple method to improve classification networks, by having a generative model synthesize difficult samples. The method is based on a regular GAN game, but includes an adversarial loss which steers the generator towards difficult samples. In comparison to previous work, we do not restrict, nor limit the kind of augmentations that the generative model can learn. We find that the generative model is able to synthesize realistically looking images which are hard to label correctly for the classification model. Since our method simply relies on backpropagation, future research can look whether the idea also applies to different tasks.

\textbf{Acknowledgement:} The work was supported by Toyota, and was  carried out at the TRACE Lab at KU Leuven (Toyota Research on Automated Cars in Europe - Leuven).

\section{Supplemental Materials}
\subsection*{Network Architectures - CURE}
\begin{table}[H]
  \caption{Discriminator}
  \begin{center}
    \begin{tabular}{r l l}
      \hline
      Operation & Features & Output size \\
       \hline
      48 x 48 input		& 3 					& 						\\
	ResNet block		& 48				& 24 x 24				\\
	ResNet block		& 96				& 12 x 12				\\
	ResNet block		& 192				& 6 x 6					\\
	ResNet block 		& 384				& 3 x 3					\\
	ReLU				& 					& 						\\
	Sum Pool          & 384				& 1 x 1					\\
	Linear				& 1					& 						\\ 
      \hline
      Filter size & 3 x 3 &  \\
      Initialization & \multicolumn{2}{l}{Xavier - $\sqrt{2}$} \\
      \hline
      \end{tabular}
    \label{tab: discriminator_cure}
  \end{center}
\end{table}

\begin{table}[H]
  \caption{Generator}
  \begin{center}
    \begin{tabular}{r l l}
      \hline
      Operation & Features & Output size \\
      \hline
      100 x 1 noise &  &  \\
      Linear        & 384   & 3 x 3 \\
      ResNet block  & 384   & 6 x 6 \\
      ResNet block  & 192   & 12 x 12 \\
      ResNet block  & 96    & 24 x 24 \\
      ResNet block  & 48    & 48 x 48 \\
      BatchNorm     &       &       \\
      ReLU          &       &       \\
      Convolution   & 3     & 48 x 48 \\
      \hline
      Filter size & 3 x 3 &  \\
      Initialization & \multicolumn{2}{l}{Xavier - $\sqrt{2}$} \\
      \hline
      \end{tabular}
    \label{tab: generator_cure}
  \end{center}
\end{table}

\begin{table}[H]
  \caption{Classifier}
  \begin{center}
    \begin{tabular}{r l l l}
      \hline
      Operation & Features & Kernel & Nonlinearity  \\
      \hline
      Convolution & 100    & 7 x 7 & ReLU \\
      Convolution & 150    & 4 x 4 & ReLU \\
      Convolution & 250    & 4 x 4 & ReLU \\
      Linear      & 300    &       & ReLU \\
      Linear      & 43     &       &    \\
      \hline
      Initialization & \multicolumn{3}{l}{Xavier - $\sqrt{2}$} \\
      \multicolumn{4}{l}{Batch normalization after each convolution} \\
      \multicolumn{4}{l}{Dropout $\left(p=0.5 \right)$ in linear layers} \\
      \hline
      \end{tabular}
    \label{tab: classifier_cure}
  \end{center}
\end{table}

\bibliographystyle{IEEEtran}
\bibliography{bib}

\begin{thebibliography}{10}
\providecommand{\url}[1]{#1}
\csname url@samestyle\endcsname
\providecommand{\newblock}{\relax}
\providecommand{\bibinfo}[2]{#2}
\providecommand{\BIBentrySTDinterwordspacing}{\spaceskip=0pt\relax}
\providecommand{\BIBentryALTinterwordstretchfactor}{4}
\providecommand{\BIBentryALTinterwordspacing}{\spaceskip=\fontdimen2\font plus
\BIBentryALTinterwordstretchfactor\fontdimen3\font minus
  \fontdimen4\font\relax}
\providecommand{\BIBforeignlanguage}[2]{{%
\expandafter\ifx\csname l@#1\endcsname\relax
\typeout{** WARNING: IEEEtran.bst: No hyphenation pattern has been}%
\typeout{** loaded for the language `#1'. Using the pattern for}%
\typeout{** the default language instead.}%
\else
\language=\csname l@#1\endcsname
\fi
#2}}
\providecommand{\BIBdecl}{\relax}
\BIBdecl

\bibitem{goodfellow2014generative}
I.~Goodfellow, J.~Pouget-Abadie, M.~Mirza, B.~Xu, D.~Warde-Farley, S.~Ozair,
  A.~Courville, and Y.~Bengio, ``Generative adversarial nets,'' in
  \emph{Advances in neural information processing systems}, 2014, pp.
  2672--2680.

\bibitem{miyato2018spectral}
T.~Miyato, T.~Kataoka, M.~Koyama, and Y.~Yoshida, ``Spectral normalization for
  generative adversarial networks,'' \emph{arXiv preprint arXiv:1802.05957},
  2018.

\bibitem{zhang2018self}
H.~Zhang, I.~Goodfellow, D.~Metaxas, and A.~Odena, ``Self-attention generative
  adversarial networks,'' \emph{arXiv preprint arXiv:1805.08318}, 2018.

\bibitem{brock2018large}
A.~Brock, J.~Donahue, and K.~Simonyan, ``Large scale gan training for high
  fidelity natural image synthesis,'' \emph{arXiv preprint arXiv:1809.11096},
  2018.

\bibitem{peng2018jointly}
X.~Peng, Z.~Tang, F.~Yang, R.~S. Feris, and D.~Metaxas, ``Jointly optimize data
  augmentation and network training: Adversarial data augmentation in human
  pose estimation,'' in \emph{Proceedings of the IEEE Conference on Computer
  Vision and Pattern Recognition}, 2018, pp. 2226--2234.

\bibitem{springenberg2015unsupervised}
J.~T. Springenberg, ``Unsupervised and semi-supervised learning with
  categorical generative adversarial networks,'' \emph{arXiv preprint
  arXiv:1511.06390}, 2015.

\bibitem{odena2016semi}
A.~Odena, ``Semi-supervised learning with generative adversarial networks,''
  \emph{arXiv preprint arXiv:1606.01583}, 2016.

\bibitem{salimans2016improved}
T.~Salimans, I.~Goodfellow, W.~Zaremba, V.~Cheung, A.~Radford, and X.~Chen,
  ``Improved techniques for training gans,'' in \emph{Advances in Neural
  Information Processing Systems}, 2016, pp. 2234--2242.

\bibitem{li2017triple}
C.~Li, K.~Xu, J.~Zhu, and B.~Zhang, ``Triple generative adversarial nets,''
  \emph{arXiv preprint arXiv:1703.02291}, 2017.

\bibitem{temel2017cure}
D.~Temel, G.~Kwon, M.~Prabhushankar, and G.~AlRegib, ``Cure-tsr: Challenging
  unreal and real environments for traffic sign recognition,'' \emph{arXiv
  preprint arXiv:1712.02463}, 2017.

\bibitem{ganin2014unsupervised}
Y.~Ganin and V.~Lempitsky, ``Unsupervised domain adaptation by
  backpropagation,'' \emph{arXiv preprint arXiv:1409.7495}, 2014.

\bibitem{goodfellow2015explaining}
I.~Goodfellow, J.~Shlens, and C.~Szegedy, ``Explaining and harnessing
  adversarial examples,'' 2015.

\bibitem{odena2017conditional}
A.~Odena, C.~Olah, and J.~Shlens, ``Conditional image synthesis with auxiliary
  classifier gans,'' in \emph{Proceedings of the 34th International Conference
  on Machine Learning-Volume 70}.\hskip 1em plus 0.5em minus 0.4em\relax JMLR.
  org, 2017, pp. 2642--2651.

\bibitem{cirecsan2012multi}
D.~Cire{\c{s}}an, U.~Meier, and J.~Schmidhuber, ``Multi-column deep neural
  networks for image classification,'' \emph{arXiv preprint arXiv:1202.2745},
  2012.

\bibitem{kingma2014adam}
D.~P. Kingma and J.~Ba, ``Adam: A method for stochastic optimization,''
  \emph{arXiv preprint arXiv:1412.6980}, 2014.

\bibitem{miyato2018cgans}
T.~Miyato and M.~Koyama, ``cgans with projection discriminator,'' \emph{arXiv
  preprint arXiv:1802.05637}, 2018.

\end{thebibliography}

\end{document}